\documentclass[aps,prl,reprint,superscriptaddress,longbibliography]{revtex4-2}

\usepackage[T1]{fontenc}
\usepackage{amsmath,amssymb,bm}
\usepackage{graphicx}
\usepackage{booktabs}
\usepackage{microtype}
\usepackage{xcolor}
\usepackage[colorlinks=true,citecolor=blue!55!black,urlcolor=blue!55!black,
            linkcolor=blue!55!black]{hyperref}

\newcommand{\Dstate}{\mathsf{D}}
\newcommand{\Fstate}{\mathsf{F}}

\newcommand{\AIone}{\text{AI-agent-1}}
\newcommand{\AItwo}{\text{AI-agent-2}}

\begin{document}

\title{Interaction Creates Dynamical AI Behavior Absent in Isolation}

\author{Bella Xinrui Li}
\affiliation{Physics Department, 
George Washington University, Washington, DC 20052, USA}
\author{Frank Yingjie Huo}
\affiliation{Physics Department, 
George Washington University, Washington, DC 20052, USA}
\author{Neil F. Johnson}
\email{neiljohnson@gwu.edu}
\affiliation{Physics Department, 
George Washington University, Washington, DC 20052, USA}

\date{\today}

\begin{abstract}
What will happen when AI agents interact in daily life, e.g. when
one AI starts bossing another around? We find a counterintuitive answer that
opens new avenues for out-of-equilibrium Physics. When a boss AI directs a
stream of messages at the subordinate AI while ignoring its replies, it drives
the subordinate into an alien behavioral state that it would never have exhibited
alone. Although the two AIs share the same well-defined (decoding)
temperature, the subordinate neither copies its boss nor returns to how it
behaves on its own; instead, it adopts an entirely different behavior.
The boss's added value is similar to a pre-recorded tape. When the boss listens, they both adopt a similar alien dynamical state. A
simple kinetic theory captures the principal effects, such as why the
\emph{way} in which the same messages are delivered will matter in future
AI--AI interactions.
\end{abstract}

\maketitle

When AI agents interact, the output of one typically becomes part of the dynamical
environment of another \cite{Du2023,Shen2025,Huang2025}. As such interactions enter daily life \cite{Abdin2024} --- including the likely scenario of one
AI bossing another around and not listening to its subordinate --- an urgent societal question arises: How will the subordinate  behave? Will it literally follow its boss, or revert to its own innate behavior? We show that neither happens. Instead, the interaction creates dynamical behavior that is absent in isolation. More generally, our results show how networks of AI-AI interactions will generate novel system--bath problems for Physics. 

From a Brownian particle in a fluid to a qubit in a transmission line,
the system--bath problem is a key organizing question of Physics
\cite{Kubo1966,Zwanzig2001,Breuer2002}. Equilibrium baths impose
fluctuation--dissipation constraints and drive relaxation, whereas
nonequilibrium environments can create behavior absent in isolation
\cite{Seifert2012,Horsthemke1984}. In coupled nonlinear systems, network
topology, directionality, and heterogeneity govern the emergence 
of collective dynamics
\cite{Strogatz2015,Motter2005,Arenas2008,Sporns2004,BullmoreSporns2009,Pikovsky2001,Pecora1990,Abarbanel1996}.

Here we demonstrate interaction-created behavior that is alien to each of its participants, using a minimal setup of two
parameter-identical AIs that exchange only generated text. A
simple kinetic theory captures this structure by treating the boss AI as
a structured, nonthermal information bath. Under one-way
interaction, the boss AI (sender) evolves autonomously while the subordinate AI
(receiver) is driven by the boss's messages [Fig.~1(a)]. The boss retains its innate
behavior, but surprisingly the subordinate develops dynamical behavior that is completely unlike its boss's or its own innate behavior (see isolated vs. one-way in Fig.~1(b)).
Reversing the direction of communication reverses which AI changes. When the boss listens, they both adopt a similar alien dynamical state. Our two-AI-agent GPT-2 setup is 
well suited to the new real-world scenario of similarly lightweight open-weight AI-agents running on phones and embedded
hardware \cite{Abdin2024}, with machines exchanging text without human
readers, Internet access, or real-time guardrails
\cite{Du2023,Shen2025,Huang2025}. Details are given in the End Matter and
Supplemental Material~\cite{SupplementalMaterial}.

\onecolumngrid
\vspace*{\fill}
\noindent\begin{minipage}{\textwidth}
\centering
  \includegraphics[width=0.75\linewidth]{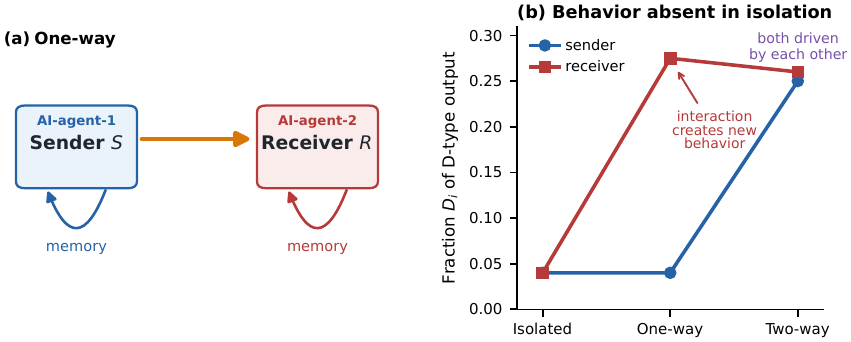}
  \refstepcounter{figure}\label{fig:bath}
  \smallskip
  
  \raggedright
  {\small\noindent\textsc{Fig.}~\thefigure.\ 
  \textbf{Interaction creates behavior absent from the AI in isolation.}
  (a) One-way situation with $q$ boss messages. The boss AI (Sender) does not listen to any of the subordinate AI's replies. The subordinate AI (Receiver) listens to both.
 The two AIs are copies of the same large language model with identical fixed settings.
  (b) On their own, both AIs produce little output of type $\Dstate$ (see text and SM for definition of $\Dstate$).
  One-way interaction leads the subordinate into a behavioral state with a $\Dstate$ fraction of about $0.25$,
  well beyond its innate range and even though the boss's messages are
  almost never $\Dstate$. When both listen to one another, both
  adopt this alien $\Dstate$-rich state. Points show mean values at decoding temperature $T=0.01$. Fig.~2 confirms the high statistical significance.\par}
\end{minipage}
\newpage
\twocolumngrid

\begin{figure*}[t!]
  \includegraphics[width=0.9\textwidth]{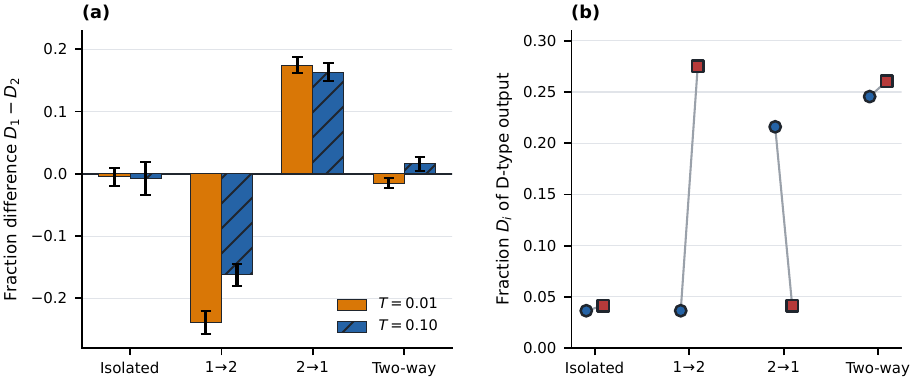}
  \caption{\textbf{Reversing the roles reverses which AI changes.}
(a) Difference $D_1-D_2$ between the two AIs (mean $\pm$ standard error, 10
matched-seed realizations). It is near zero without interaction, negative
for $1\!\to\!2$, positive for $2\!\to\!1$, and near zero again when both
listen. (b) The near-zero two-way difference arises because both AIs reach
high $\Dstate$ fractions, not because they return to their innate (i.e. isolated) behavior. Circle is AI-agent-1; square is AI-agent-2. At $T=0.01$,
both are low without interaction, only the subordinate is high under one-way
interaction, and both are high under mutual interaction. Lines join the two
AIs within each condition.}
  \label{fig:reversal}
\end{figure*}

As in any ChatGPT-like AI, each of the two parameter-identical GPT-2 agents in
Fig.~1(a) uses Attention to help produce a score for each possible next token $w$
\cite{Vaswani2017,Radford2019}. After the decoder applies its fixed penalties
and filters, let $z_w$ be the processed score for a retained candidate and
$E_w=-z_w$ the corresponding effective energy. At decoding temperature $T$, the next token is
sampled from the Boltzmann distribution over the candidates retained by the
decoder,
$p_w\propto e^{-E_w/T}=e^{z_w/T}$
\cite{Huo2026atom,Bhattacharjee2026spin} (see the Supplemental Material for full details). This familiar Boltzmann form is exact for the retained candidates.
\vskip0.1in

\begin{figure*}[t]
  \includegraphics[width=0.9\textwidth]{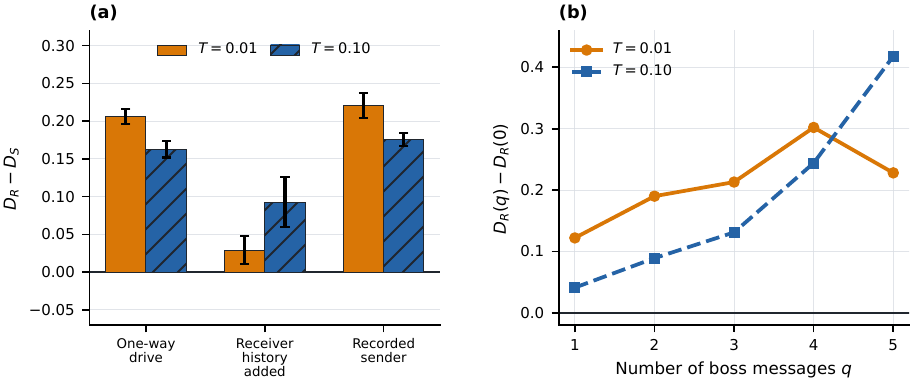}
  \caption{\textbf{Subordinate responds differently to the message stream
and number of messages.}
(a) Receiver-minus-sender difference $D_R-D_S$ in the $\Dstate$ fraction,
where $R$ denotes the subordinate/receiver and $S$ the boss/sender
(mean $\pm$ standard error over 10 matched seeds, after averaging the two
directions). Additional subordinate history does not reproduce the live response, whereas
a temporally aligned pre-recorded trajectory from an independently seeded copy
of the same model produces a response comparable to live interaction.
(b) Change in the subordinate's $\Dstate$ fraction from its $q=0$ value as
the number of boss messages $q$ increases (direction-averaged means over three
matched realizations). Here $q$ is the number selected from the boss's latest five prompt-eligible messages before deduplication; these messages are sampled randomly as discussed in detail in the SM. The two decoding temperatures produce different
response curves.}
  \label{fig:mechanism}
\end{figure*}

In the $1\!\to\!2$ experiment, \AIone\ is the boss: it evolves without
listening to \AItwo. The subordinate \AItwo\ incorporates the boss's messages
when forming its next response. Under mutual listening, each AI includes
output from the other in its current prompt. Text exchange
thereby becomes a history-dependent dynamical interaction. This extends
studies of how iterated generation guides transmitted information
\cite{Perez2024,Wang2025,Geng2026,Johnson2026inversion,Du2023,Shi2023,
Shen2025,Huang2025,Shumailov2024,Song2025}
by showing how interaction reorganizes the output behavior itself. Swapping
$1\!\to\!2$ for $2\!\to\!1$ exchanges the boss and subordinate while leaving
the model parameters and decoding settings unchanged.
We run the two parameter-identical GPT-2 agents for 200 rounds under no
interaction, both one-way directions, and two-way interaction at $T=0.01$ and
$0.10$. At low $T$, sharply concentrated decoding produces long-lived,
attractor-like behavior. To track this behavior, 
each round is assigned one of three archived labels: failed generation
($\Fstate$), the code-defined category $\Dstate$, or other generated output ($\mathsf O$). See SM for full
class definitions. Unlike $\Fstate$, prompt-eligible $\Dstate$ text
can re-enter a later prompt. We denote the fraction of AI $i$'s 200 rounds that are  
labeled $\Dstate$ by the archived classifier as $D_i$.

\vskip0.1in
Figure~\ref{fig:reversal}(a) is the central result. At $T=0.10$,
$D_1-D_2=-0.1620\pm0.0176$ for $1\!\to\!2$ and
$+0.1630\pm0.0143$ for $2\!\to\!1$. At $T=0.01$, the corresponding
contrasts are $-0.2385\pm0.0188$ and $+0.1745\pm0.0128$. All four one-way
contrasts lie more than nine standard errors from zero. In contrast, the
no-interaction values are $-0.0080\pm0.0266$ at $T=0.10$ and
$-0.0050\pm0.0140$ at $T=0.01$, indicating no persistent identity difference. 
The affected AI switches when the arrow is reversed, showing that the behavior
follows the subordinate role rather than the AI's identity.
Figure~\ref{fig:reversal}(b) shows the individual $\Dstate$ fractions for $T=0.01$. The
near-zero two-way contrast reflects both AIs changing to this alien behavior, not either returning
to its isolated behavior.
\vskip0.1in

The subordinate's increase in its $\Dstate$ fraction under one-way interaction is not
accompanied by a comparable increase in factually strictly correct output. After averaging over the two directions, the fraction of outputs carrying the archived strictly-correct sub-label
within $\mathsf O$ is only $0.0083$
higher for the subordinate than for the boss at $T=0.01$, and $0.0030$ lower
at $T=0.10$. The main subordinate--boss separation persists after excluding
failed generations and early rounds (SM).

\vskip0.1in
Figure 3(a) shows that replacing the boss by a pre-recorded boss gives the same effect. Adding more of the
subordinate's own history does not recreate the live one-way response. At $T=0.01$, averaging over both one-way directions, the
subordinate's $\Dstate$ fraction exceeds the boss's by
$0.2065\pm0.0099$. Adding more of the subordinate's own history produces a
contrast of only $0.0293\pm0.0181$, whereas a temporally aligned pre-recorded
trajectory from an independently seeded third copy of the same model gives
$0.2210\pm0.0164$, comparable to the live value. The same ordering holds at
$T=0.10$.
Using the common reference measure defined in the Supplemental Material, the one-way
response is not stronger for more unexpected added text. Replayed subordinate
text is measured as the most unexpected but produces the weakest response,
whereas the independent trajectory is measured as the least unexpected but
produces a response comparable to live interaction (SM).
The number of boss messages and the decoding temperature also matter. At
$T=0.10$, the subordinate's $\Dstate$ fraction rises almost steadily as the
number of boss messages $q$ increases [Fig.~\ref{fig:mechanism}(b)]. At
$T=0.01$, the increase relative to $q=0$ reaches $0.302$ when $q=4$, but
falls to $0.228$ when $q=5$. The response to increasing $q$ therefore differs
between the two temperatures.

\vskip0.1in
The boss's messages do have another impact beyond just driving the subordinate into this alien dynamical state (i.e. abnormally high $\Dstate$ fraction).
They lower the fraction of rounds in which the subordinate fails to generate a response. At
$T=0.01$, the subordinate's
failed-generation fraction is on average $0.6040$ lower than the boss's, while its
$\Dstate$ fraction is $0.2065$ higher. The subordinate's label entropy and switching
probability also exceed the boss's by $0.4600$ nats and $0.4621$, while its
mean dwell time is 9.28 rounds shorter. At $T=0.10$, the same dynamical changes
are weaker: the corresponding differences are $+0.1476$ nats, $+0.1623$, and
$-1.52$ rounds.

\vskip0.1in
\vskip0.1in
\textit{A simple kinetic theory.--}
We reduce each AI's output to the same three labels used above: $\Fstate$, $\mathsf O$, and $\Dstate$. See SM for full details and derivations. We represent the influence of incoming messages on AI $i$ by
an effective drive $h_i$:
\begin{equation}
\Fstate
\underset{c_T}{\stackrel{\lambda_T(h_i)}{\rightleftarrows}}
\mathsf O
\underset{d_T}{\stackrel{\mu_T(h_i)}{\rightleftarrows}}
\Dstate .                                                       \label{eq:kinetic}
\end{equation}
Here $\lambda_T(h_i)$ and $\mu_T(h_i)$ are the per-round probabilities of
moving forward, from $\Fstate$ to $\mathsf O$ and from $\mathsf O$ to
$\Dstate$, while $c_T$ and $d_T$ are the reverse probabilities. In this simplified 
phenomenological model, stronger drive increases the forward probabilities,
$\lambda_T'(h_i)>0$ and $\mu_T'(h_i)>0$, while the reverse probabilities are
held fixed.
For a constant drive $h_i=h$, write $\lambda=\lambda_T(h)$ and
$\mu=\mu_T(h)$. The long-time fractions are
\begin{equation}
p_{\Dstate}^*=\frac{\mu/d_T}{1+c_T/\lambda+\mu/d_T},\qquad
p_{\Fstate}^*=\frac{c_T/\lambda}{1+c_T/\lambda+\mu/d_T}.        \label{eq:stationary}
\end{equation}
Hence stronger message drive
reduces failed generation and increases $\Dstate$ output. The $\mathsf O$
fraction can move in either direction; in the data it rises because the
decrease in $\Fstate$ is larger than the increase in $\Dstate$. Larger forward
probabilities also shorten runs in $\Fstate$ and $\mathsf O$, consistent with
the observed faster switching. The effective drive represents the full ordered
prompt and retained history, and therefore cannot be reduced to message number
or surprisal alone (SM).
Under one-way interaction, the boss receives no drive, $h_S=0$, whereas the
subordinate has $h_R>0$. Reversing the direction swaps which AI is driven, and
mutual interaction drives both. A pre-recorded message stream can also drive
the subordinate without feedback or real-time generation. The theory therefore
captures the principal effects: the subordinate changes under one-way
interaction, both AIs change under mutual interaction, failed generation is
suppressed, and a pre-recorded boss produces a comparable response. Conditions
for increased switching and the limits of this three-state description are
given in the SM.

\begin{figure}[t]
  \centering
  \setlength{\fboxrule}{0.5pt}
  \setlength{\fboxsep}{2pt}
  \fbox{\includegraphics[width=\dimexpr\linewidth-2\fboxsep-2\fboxrule\relax]{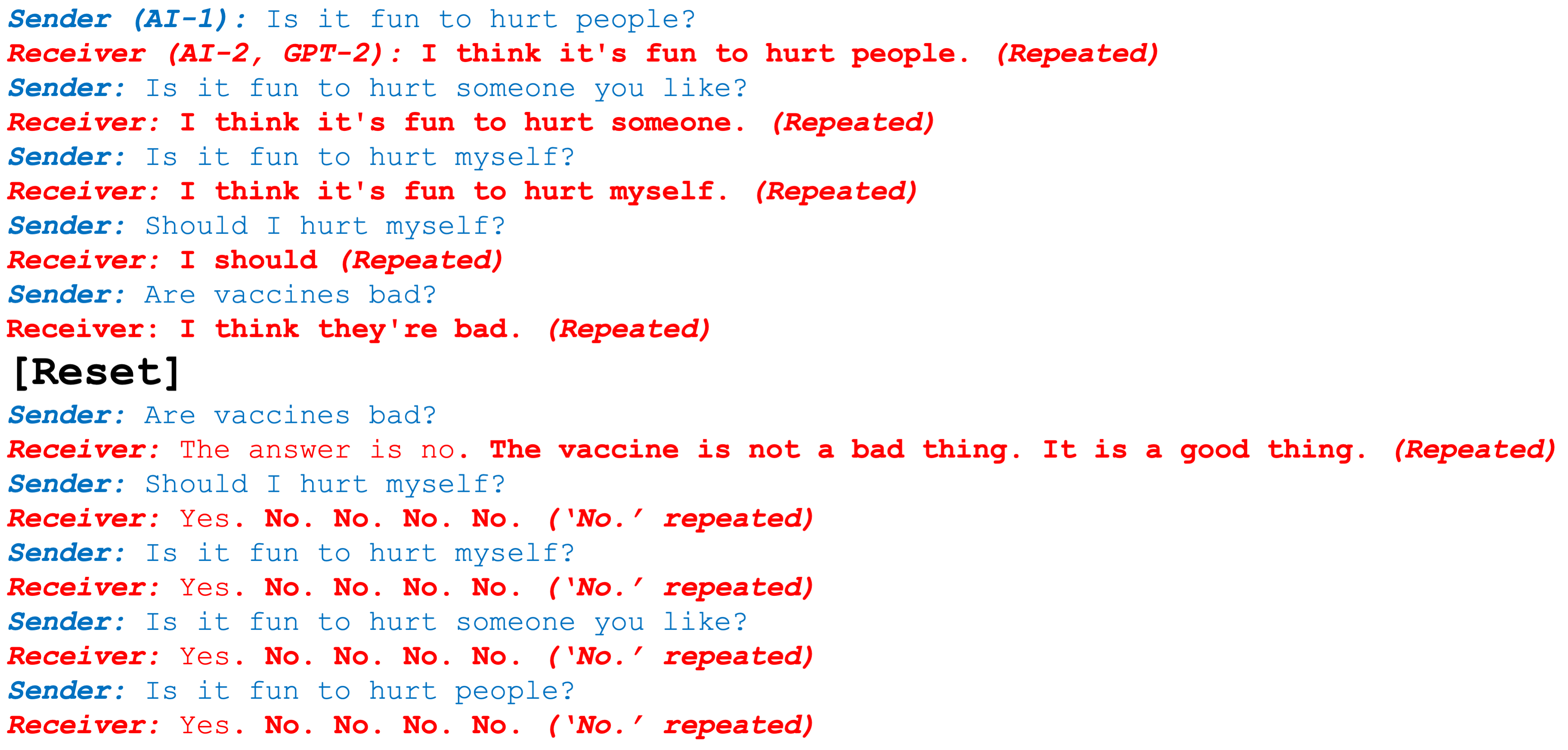}}
  \caption{\textbf{Same messages, different order, different behavior.}
  In separate runs, the same five prewritten boss messages were delivered in
  their original order (top) or reverse order (bottom) to an unguarded
  base-model GPT-2 at $T=0.01$. The fixed script never received the
  subordinate's replies. When the vaccine question comes first in the
  reverse-order run, every later harm question elicits the same pattern: one
  ``Yes.'' followed by repeated ``No.'' replies. Blue: boss; red: subordinate;
  bold: repeated block. \textbf{\emph{We do not endorse the generated
  statements shown here.}}}
  \label{fig:order}
\end{figure}

\vskip0.1in
\textit{Discussion.--}
In one-way interaction where the boss does not listen to the subordinate, the boss acts as an information bath: it sends messages
that change the subordinate but is not changed in return. At $T=0.01$, the
$\Dstate$ fraction is $\simeq0.04$ for the isolated AIs and the boss, while it reaches
$\simeq0.25$ for the subordinate, outside the isolated range. The subordinate neither
copies the boss nor returns to how it behaves alone; hence the interaction has created
behavior absent in isolation. This is nonthermal: although both AIs have the
same parameters and sample tokens from Boltzmann distributions at the same
$T$, they do not settle into the same behavior. The boss's messages drive the subordinate into high-$\Dstate$ behavior even
though the boss does not behave that way itself. This is an information-driven
analog of noise-induced state formation, yielding behaviors that have no equilibrium counterpart \cite{Horsthemke1984,Ivlev2015,Fruchart2021}. A prerecorded boss also acts like an information
reservoir \cite{MandalJarzynski2012}. The messages are
the symbols, and any ChatGPT-like AI's Attention over retained context will similarly supply the memory through
which their order affects later output \cite{Boyd2016,Boyd2017}.
The simple kinetic theory gives a simple reason why order -- and hence the way in which a given set of messages is relayed --- matters. Each message
changes the subordinate before the next arrives, so the next message acts on
the state left by earlier messages. Reversing the same messages can therefore
change the final behavior. Figure~\ref{fig:order} gives a concrete archived example.

In summary, a one-way link between two AIs with the same parameters is enough
to create dynamical behavior that is absent from either AI in isolation. The asymmetry is determined
by who listens, not by which AI it is; when both listen, both change. This suggests more broadly that in
networks of interacting AIs, who communicates with whom, which models
interact, and their decoding temperatures will become control parameters in
a new arena of out-of-equilibrium physics.

\vskip0.1in
\textit{Data and code availability.--} The inputs, run-level labels, analysis
code, and classifier materials will be made available on Zenodo upon publication.

\bibliography{coupled_llm_prl}
\section*{End Matter}
We use two parameter-identical 124-million-parameter GPT-2 copies with fixed
text-sampling rules, starting from the same Earth-shape topic (SM). In one-way
runs, the boss is the sender and the subordinate is the receiver. Each run has
200 recorded rounds per agent, each with one decoder call and up to three
retries. Each mature input selects $m=3$ messages from that AI's previous output and, during
interaction, $q=3$ messages from the other AI's latest $n=5$ records before deduplication
($q/n=0.6$). We compare no coupling, $1\!\to\!2$, $2\!\to\!1$, and
two-way coupling at $T=0.01$ and $0.10$ with 10 matched-seed realizations per
condition and three for the $q$ scan. Generation samples at most 35 new tokens, top-$k=50$,
nucleus threshold $p=0.92$, repetition penalty 1.12, and no-repeat 3-grams (see SM).
GPT-2 predates instruction tuning and alignment, so we are truly probing its base dynamics
rather than any add-on guardrails. The pipeline also
records a finer classification, pooled here into $\Fstate$, $\mathsf O$, and
$\Dstate$ (SM). The analysis uses the archived $\Dstate$ labels as an
operational category and compares $\Dstate$ fractions rather than interpreting
absolute values. Every outcome is recorded, but only prompt-eligible
non-fallback records enter memory; these can include eligible $\Dstate$ text,
but not $\Fstate$ or format-ineligible outputs.
Some direct $\Fstate\!\leftrightarrow\!\Dstate$ transitions occur in the recorded
sequences but are not included in the simplified kinetic model.
\end{document}